%% file: paper.tex
\documentclass[runningheads,a4paper]{llncs}

\usepackage{amssymb}
\usepackage{amsmath}
\usepackage{graphicx}
\usepackage[table,pdftex]{xcolor}
\usepackage{xspace}
\usepackage[colorlinks=true,citecolor=blue,urlcolor=black]{hyperref}
\usepackage{lineno}
\usepackage{subfig}
\usepackage[colorinlistoftodos]{todonotes}
\usepackage{wrapfig}
\usepackage[misc]{ifsym}
\usepackage{amsfonts}
\usepackage{booktabs}
\usepackage{siunitx}

\begin{document}

\mainmatter  
\title{Deep nested level sets: Fully automated segmentation of cardiac MR images in patients with pulmonary hypertension}
\titlerunning{Deep nested level set method}

\newcommand{\corrauth}{\textsuperscript{(\Letter)}}
\author{Jinming Duan$^{1,2}$\corrauth, Jo Schlemper$^{1}$, Wenjia Bai$^{1}$, Timothy J W Dawes$^{2}$, Ghalib Bello$^{2}$, Georgia Doumou$^{2}$, Antonio De Marvao$^{2}$, Declan P O'Regan$^{2}$, Daniel Rueckert$^{1}$}
\authorrunning{J Duan, et al.}
\institute{$^1$Biomedical Image Analysis Group, Imperial College London, London, UK\\
$^2$MRC London Institute of Medical Sciences, Imperial College London, London, UK\\
\email{j.duan@imperial.ac.uk}\\}

\maketitle


\input{sections/abstract}
\input{sections/introduction}
\input{sections/AK}
\input{sections/method}
\input{sections/results}

\input{sections/conclusion}
\subsubsection*{Acknowledgements.} 
The research was supported by the British Heart Foundation (NH/17/1/32725); National Institute for Health Research (NIHR) Biomedical Research Centre based at Imperial College Healthcare NHS Trust and Imperial College London; and the Medical Research Council, UK. We would like to thank Dr Simon Gibbs, Dr Luke Howard and Prof Martin Wilkins for providing the CMR image data. The Titan Xp for this research was donated by NVIDIA. 

\bibliographystyle{splncs}
\bibliography{cites}

\end{document}

%% file: sections/abstract.tex

\begin{abstract}

In this paper we introduce a novel and accurate optimisation method for segmentation of cardiac MR (CMR) images in patients with pulmonary hypertension (PH). The proposed method explicitly takes into account the image features learned from a deep neural network. To this end, we estimate simultaneous probability maps over region and edge locations in CMR images using a fully convolutional network. Due to the distinct morphology of the heart in patients with PH, these probability maps can then be incorporated in a single nested level set optimisation framework to achieve multi-region segmentation with high efficiency. The proposed method uses an automatic way for level set initialisation and thus the whole optimisation is fully automated. We demonstrate that the proposed deep nested level set (DNLS) method outperforms existing state-of-the-art methods for CMR segmentation in PH patients.

\end{abstract}

%% file: sections/introduction.tex

\section{Introduction}

Pulmonary hypertension (PH) is a cardiorespiratory syndrome characterised by increased blood pressure in pulmonary arteries. It typically follows a rapidly progressive course. As such, early identification of PH patients with elevated risk of a deteriorating course is of paramount importance. For this, accurate segmentation of different functional regions of the heart in CMR images is critical.  

Numerous methods for automatic and semi-automatic CMR image segmentation have been proposed, including deformable models \cite{feng2013segmentation}, atlas-based image registration models \cite{bai2013probabilistic} as well as statistical shape and appearance models \cite{alba2016algorithm}. More recently, deep learning-based methods have achieved state-of-the-art performance in the CMR domain \cite{bai2017human}. However, the above approaches for CMR image segmentation have multiple drawbacks. First, they tend to focus on left ventricle (LV) \cite{feng2013segmentation}. However, the prognostic importance of the right ventricle (RV) is a broad range of cardiovascular disease and using the coupled biventricular motion of the heart enables more accurate cardiac assessment. Second, existing approaches rely on manual initialisation of the image segmentation or definition of key anatomical landmarks \cite{feng2013segmentation,bai2013probabilistic,alba2016algorithm}. This becomes less feasible in population-level applications involving hundreds or thousands of CMR images. Third, existing techniques have been mainly developed and validated using normal (healthy) hearts \cite{feng2013segmentation,bai2013probabilistic,bai2017human}. Few studies have focused on abnormal hearts in PH patients.

To address the aforementioned limitations of current approaches, in this paper we propose a deep nested level set (DNLS) method for automated biventricular segmentation of CMR images. More specifically, we make three distinct contributions to the area of CMR segmentation, particularly for PH patients: First, we introduce a deep fully convolutional network that effectively combines two loss functions, i.e. softmax cross-entropy and class-balanced sigmoid cross-entropy. As such, the neural network is able to simultaneously extract robust region and edge features from CMR images. Second, we introduce a novel implicit representation of PH hearts that utilises multiple nested level lines of a continuous level set function. This nested level set representation can be effectively deployed with the learned deep features from the proposed network. Furthermore, an initialisation of the level set function can be readily derived from the learned feature. Therefore, DNLS does not need user intervention (manual initialisation or landmark placement) and is fully automated. Finally, we apply the proposed DNLS method to clinical data acquired from 430 PH patients (approx. 12000 images), and compare its performance with state-of-the-art approaches.


%% file: sections/AK.tex
\section{Modelling biventricular anatomy in patients with PH}
\label{sec:method}
\vspace{-1pt}
\begin{wrapfigure}{ro}{0.4\textwidth}
\vspace{-22pt}
\centering
\includegraphics[width=0.4\textwidth]{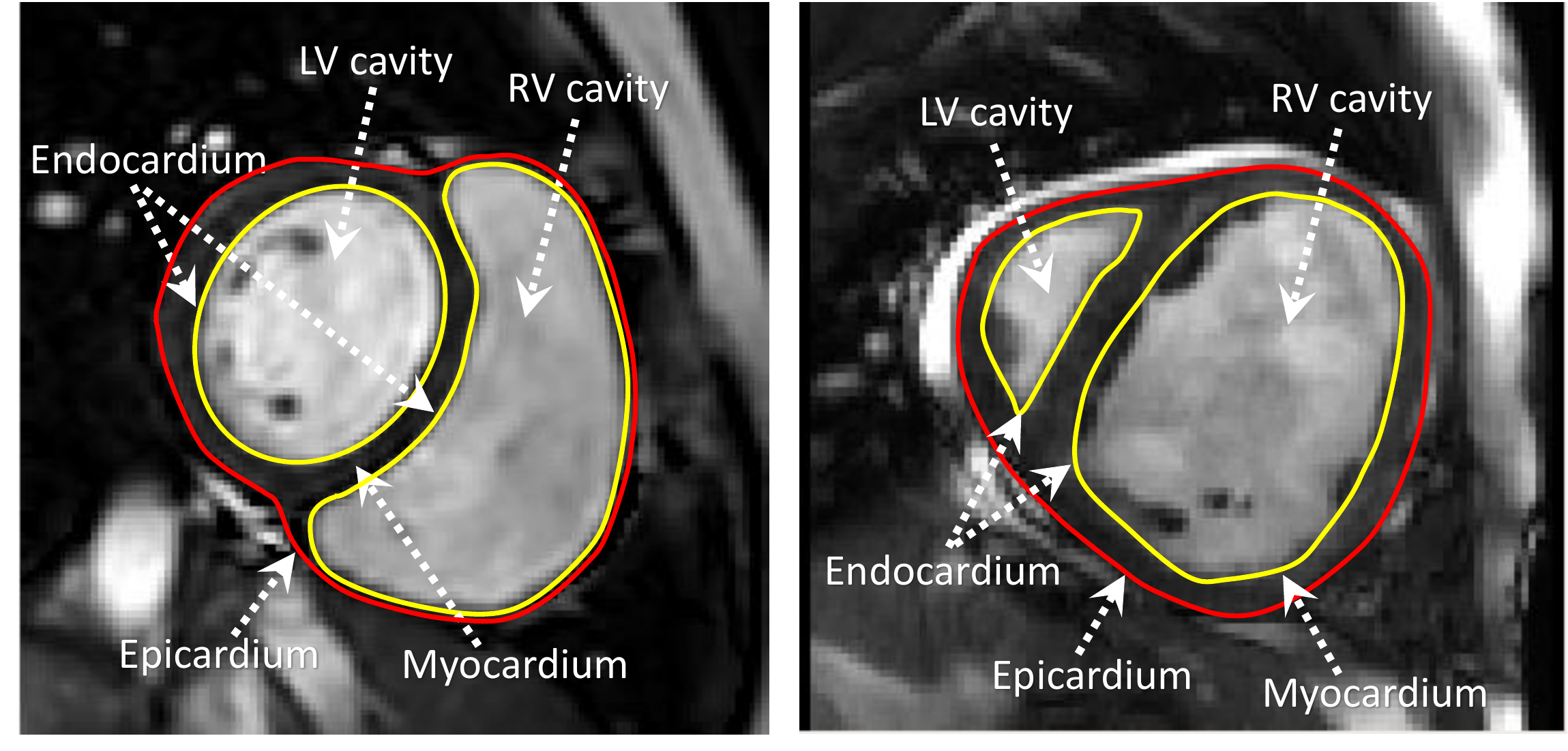}
\vspace{-20pt}
\caption{Short-axis images of a healthy subject (left) and a PH subject (right), including the anatomical explanation of both LV and RV. The desired epicardial contours (red) and endocardial contours (yellow) from both ventricles are plotted.}
\vspace{-22pt}
\label{fig:HPH}
\end{wrapfigure}
To illustrate cardiac morphology in patients with PH, Fig~\ref{fig:HPH} shows the difference in CMR images from a representative healthy subject and a PH subject. In health, the RV is crescentic in short-axis views and triangular in long-axis views, wrapping around the thicker-walled LV. In PH, the initial hypertrophic response of the RV increases contractility but is followed invariably by progressive dilatation and failure heralding clinical deterioration and ultimately death. During this deterioration, the dilated RV pushes onto the LV to deform and lose its roundness. Moreover, in PH the myocardium around RV become much thicker than a healthy one, allowing PH cardiac morphology to be modelled by a nested level set. Next, we incorporate the biventricular anatomy of PH hearts into our model for automated segmentation of LV and RV cavities and myocardium.

%% file: sections/method.tex

\section{Methodology}
\label{sec:method}

\textbf{Nested level set approach:} We view image segmentation in PH as a multi-region image segmentation problem. Let $I : \Omega \to {\mathbb{R}^d}$ denote an input image defined on the domain $\Omega \subset \mathbb{R}^2$. We segment the image into a set of $n$ pairwise disjoint region $\Omega_i$, with $\Omega  =  \cup _{i = 1}^n{\Omega _i}$, ${\Omega _i} \cap {\Omega _j} = \emptyset$ $\forall i \ne j$. The segmentation task can be solved by computing a labelling function $l(x):\Omega \to \{1,...,n\}$ that indicates which of the $n$ regions each pixel belongs to: ${\Omega _i} = \left\{ {x\left| {l\left( x \right) = i} \right.} \right\}$. The problem is then formulated as an energy minimisation problem consisting of a data term and a regularisation term
\begin{equation} \label{eq:MP}
\mathop {\min }\limits_{{\Omega _1},...,{\Omega _n}} \left\{ {\sum\limits_{i = 1}^n {\int_{{\Omega _i}} {{f_i}\left( x \right)dx} }  + \lambda \sum\limits_{i = 1}^n {{\rm{Pe}}{{\rm{r}}_g}\left( {{\Omega _i},\Omega } \right)} } \right\}.
\end{equation}

\begin{wrapfigure}{ro}{0.4\textwidth}
\vspace{-30pt}
\centering
\includegraphics[width=0.4\textwidth]{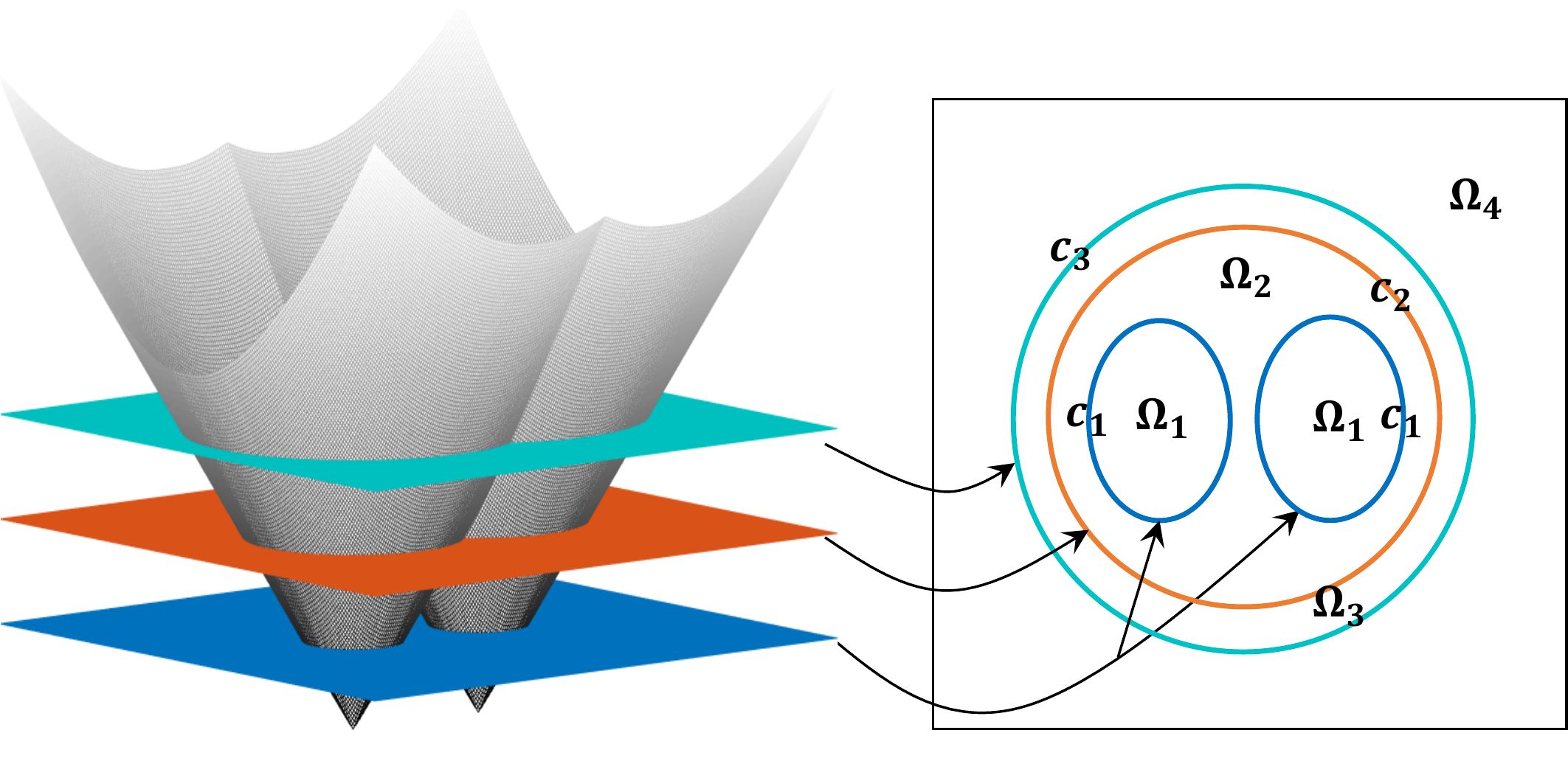}
\vspace{-15pt}
\caption{An example of partitioning the domain $\Omega$ into $4$ disjoint regions (right), using $3$ nested level lines $\{x|\phi(x)= c_i , i=1,2,3\} $ of the same function $\phi$ (left). The intersactions between the 3D smooth surface $\phi$ and the 2D plans correspond to the three nested curves on the right.}
\vspace{-22pt}
\label{fig:nestedLS}
\end{wrapfigure}

\noindent The data term, $f_i:\Omega \to \mathbb{R}$ is associated with region that takes on smaller values if the respective pixel position has stronger response to region. In a Bayesian MAP inference framework, ${f_i}\left( x \right) =  - \log P_i \left( {I\left( x \right)|{\Omega _i}} \right)$ corresponds to the negative logarithm of the conditional probability for a specific pixel color at the given location $x$ within region $\Omega_i$. Here we refer to $f_i$ as region feature. The second term, ${{\rm{Pe}}{{\rm{r}}_g}\left( {{\Omega _i},\Omega } \right)}$ is the perimeter of the segmentation region $\Omega_i$, weighted by the non-negative function $g$. This energy term alone is known as geodesic distance, the minimisation over which can be interpreted as finding a geodesic curve in a Riemannian space. The choice of $g$ can be an edge detection function which favours boundaries that have strong gradients of the input image $I$. Here we refer to $g$ as edge feature. 

We apply the variational level set method \cite{duan2014some,tan2018image} to (\ref{eq:MP}) in this study. Because a PH heart can be implicitly represented by two nested level lines of a continuous level set function ($\{x|\phi(x)=c_i, i=1,2\}$ in Fig~\ref{fig:nestedLS}). Note that the nested level set idea present here is inspired from previous work \cite{feng2013segmentation,chung2009image}. Our approach uses features learned from many images while previous work only consider single image. With the idea, we are able to approximate the multi-region segmentation energy (\ref{eq:MP}) by using only one continuous function. The computational cost is thus small. Now assume that the contours in the image $I$ can be represented by level lines of the same Lipschitz continuous level set function $\phi:\Omega \to \mathbb{R}$. With $n-1$ distinct levels $\{c_1 < c_2 < \cdot \cdot \cdot <c_{n-1}\}$, the implicit function $\phi$ partitions the domain $\Omega$ into $n$ disjoint regions, together with their boundaries (see Fig~\ref{fig:nestedLS} right). We can then define the characteristic function $\chi_i\phi$ for each region $\Omega_i$ as

\begin{equation} \label{eq:CharacticeFunction}
{\chi _i}\phi(x)  = \left\{ \begin{array}{lc}
H\left( {{c_i} - \phi(x) } \right) \;\;\; &i=1\\
H\left( {\phi(x)  - {c_{i - 1}}} \right)H\left( {{c_i} - \phi(x) } \right) \;\;\; & 2 \le i \le n-1\\
H\left( {\phi(x)  - {c_{i - 1}}} \right) \;\;\; &i=n
\end{array} \right.,
\end{equation}
where $H$ is the one-dimensional Heaviside function that takes on either 0 or 1 over the whole domain $\Omega$. Due to the non-differentiate nature of $H$ it is usually approximated by its smooth version $H_\epsilon$ for numerical calculation \cite{chung2009image}. Note that in (\ref{eq:CharacticeFunction}) $\sum\nolimits_{i = 1}^n {{\chi _i}\phi = 1}$ is automatically satisfied, meaning that the resulting segmentation will not produce a vacuum or an overlap effect. That is, by using (\ref{eq:CharacticeFunction}) $\Omega  =  \cup _{i = 1}^n{\Omega _i}$ and ${\Omega _i} \cap {\Omega _j} = \emptyset$ hold all the time. With the definition of ${\chi _i}\phi$, we can readily reformulate (\ref{eq:MP}) in the following new energy minimisation problem 
\begin{equation} \label{eq:transformedFunction}
\mathop {\min }\limits_{\phi(x)}  \left\{ {\sum\limits_{i = 1}^n {\int_\Omega  {{f_i}\left( x \right){\chi _i}\phi \left( x \right)dx} }  + \lambda \sum\limits_{i = 1}^{n - 1} {\int_\Omega  {g\left( x \right)\left| {\nabla H\left( {\phi \left( x \right) - {c_{i }}} \right)} \right|dx} } } \right\}.
\end{equation}
Note that (\ref{eq:transformedFunction}) differs from (\ref{eq:MP}) in multiple ways due to the use of the smooth function $\phi$ and characteristic function (\ref{eq:CharacticeFunction}). First, the variable to be minimised is the $n$ regions ${\Omega _1},...,{\Omega _n}$ in (\ref{eq:MP}) while the smooth function $\phi$ in (\ref{eq:transformedFunction}). Second, the minimisation domain is changing from over $\Omega_i$ in (\ref{eq:MP}) to over $\Omega$ in (\ref{eq:transformedFunction}). Third, (\ref{eq:MP}) uses an abstract ${{\rm{Pe}}{{\rm{r}}_g}\left( {{\Omega _i},\Omega } \right)}$ for the weighted length of the boundary between two adjacent regions, while (\ref{eq:transformedFunction}) represents the weighted length with the co-area formula, i.e. ${\int_\Omega  {g\left| {\nabla H\left( {\phi - {c_i}} \right)} \right|dx} }$. Finally, the upper limit of summation in the regularisation term of (\ref{eq:MP}) is $n$ while $n-1$ in that of (\ref{eq:transformedFunction}). So far, the region features $f_i$ and the edge feature $g$ have not been defined. Next, we will tackle this problem.\\


\noindent \textbf{Learning deep features using fully convolutional network:} We propose a deep neural network that can effectively learn region and edge features from many labelled PH CMR images. Learned features are then incorporated to (\ref{eq:transformedFunction}). Let us formulate the learning problem as follows: we denote the input training data set by $S=\{(U_p, R_p, E_p), p=1,...,N\}$, where sample $U_p=\{u^p_j,j=1,...,|U_p|\}$ is the raw input image, $R_p=\{r^p_j,j=1,...,|R_p|\}$, $r^p_j \in \{1,...,n\}$ is the ground truth region labels ($n$ regions) for image $U_p$, and $E_p=\{e^p_j,j=1,...,|E_p|\}$, $e^p_j \in \{0,1\}$ is the ground truth binary edge map for $U_p$. We denote all network layer parameters as $\textbf{W}$ and  propose to minimise the following objective function via the (back-propagation) stochastic gradient descent
\begin{equation} \label{eq:REloss}
\textbf{W}^* = {\rm{argmin}} (L_R(\textbf{W}) + \alpha L_E(\textbf{W}) ),
\end{equation}
where $L_R(\textbf{W})$ is the region associated cross-entropy loss that enables the network to learn region features, while $L_E(\textbf{W})$ is the edge associated cross-entropy loss for learning edge features. The weight $\alpha$ balances the two losses. By minimising (\ref{eq:REloss}), the network is able to output joint region and edge probability maps simultaneously. In our image-to-image training, the loss function is computed over all pixels in a training image $U=\{u_j,j=1,...,|U|\}$, a region map $R=\{r_j,j=1,...,|R|\}$, $r_j \in \{1,...,n\}$ and an edge map $E=\{e_j,j=1,...,|E|\}$, $e_j \in \{0,1\}$. The definitions of $L_R(\textbf{W})$ and $L_E(\textbf{W})$ are given as follows.
\begin{equation} \label{eq:Rloss}
L_R(\textbf{W}) = -\sum\limits_{j}{\rm{log}}P_{so}(r_j|U,\textbf{W}),
\end{equation} 
where $j$ denotes the pixel index, and $P_{so}(r_j|U,\textbf{W})$ is the channel-wise softmax probability provided by the network at pixel $j$ for image $U$. The edge loss is 
\begin{equation} \label{eq:Eloss}
L_E(\textbf{W}) = -\beta\sum\limits_{j \in Y_+}{\rm{log}} P_{si}(e_j=1|U,\textbf{W}) - (1-\beta)\sum\limits_{j \in Y_-}{\rm{log}}P_{si}(e_j=0|U,\textbf{W}).
\end{equation}
For a typical CMR image, the distribution of edge and non-edge pixels is heavily biased. Therefore, we use the strategy  \cite{xie2015holistically} to automatically balance edge and non-edge classes. Specifically, we use a class-balancing weight $\beta$. Here, $\beta  =  |Y_ -|/|Y|$ and $1-\beta=|Y_ +|/|Y|$, where $|Y_-|$ and $|Y_+|$ respectively denote edge and non-edge ground truth label pixels. $P_{si}(e_j=1|U,\textbf{W})$ is the pixel-wise sigmoid probability provided by the network at non-edge pixel $j$ for image $U$.

\begin{figure}[h!] 
\vspace{-10pt}
\centering  
{\includegraphics[width=0.99\textwidth]{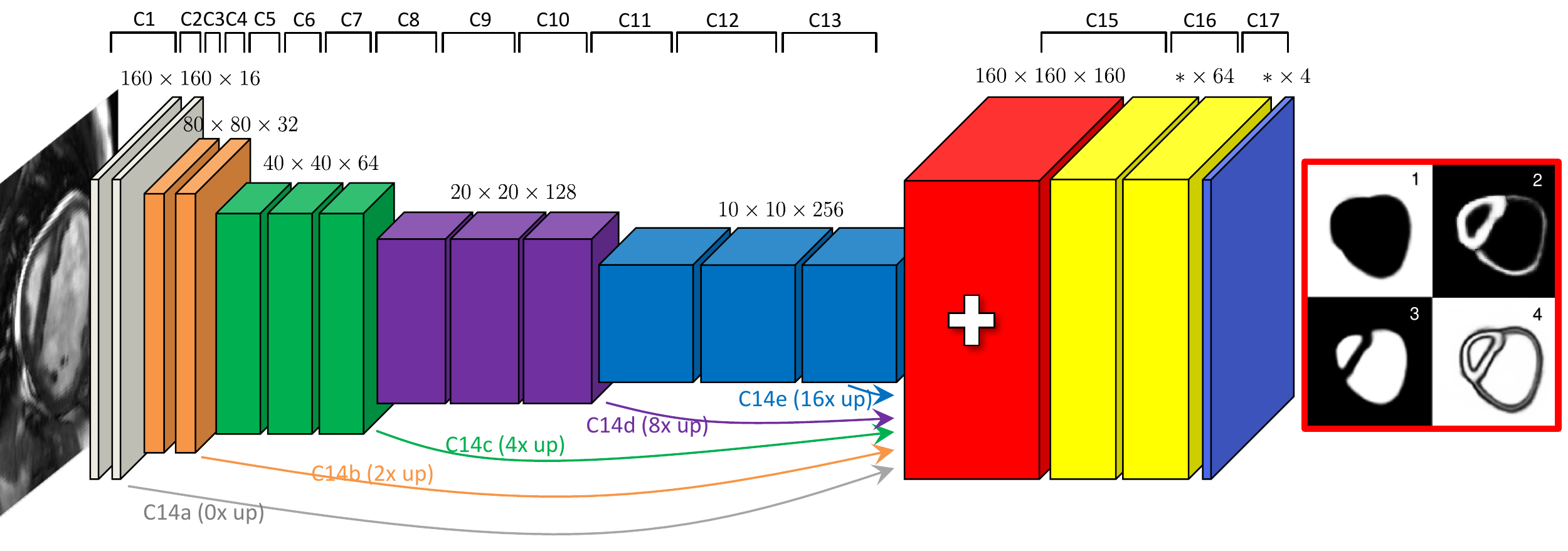}}\\
\caption{The architecture of a fully convolutional network with 17 convolutional layers. The network takes the PH CMR image as input, applies a branch of convolutions, learns image features from fine to coarse levels, concatenates (`+' sign in the red layer) multi-scale features and finally predicts the region (1-3) and edge (4) probability maps simultaneously. }
\vspace{-10pt}
\label{fig:network}
\end{figure}

In Fig~\ref{fig:network}, we show the network architecture for automatic feature extraction, which is a fully convolutional network (FCN) and adapted from the U-net architecture \cite{simonyan2014very}. Batch-normalisation (BN) is used after each convolutional layer, and before a rectified linear unit (ReLU) activation. The last layer is however followed by the softmax and sigmoid functions. In the FCN, input images have pixel dimensions of $160 \times 160$. Every layer whose label is prefixed with `C' performs the operation: convolution $\to$ BN $\to$ ReLU, except C17. The (filter size/stride) is (3$\times$3/1) for layers from C1 to C16, excluding layers C3, C5, C8 and C11 which are (3$\times$3/2). The arrows represent (3$\times$3/1) convolutional layers (C14a$-$e) followed by a transpose convolutional (up) layer with a factor necessary to achieve feature map volumes with size 160 $\times$ 160 $\times$ 32, all of which are concatenated into the red feature map volume. Finally, C17 applies a (1$\times$1/1) convolution with a softmax activation and a sigmoid activation, producing the blue feature map volume with a depth $n+1$, corresponding to $n$ (3) region features and an edge feature of an image.

After the network is trained, we deploy it on the given image $I$ in the validation set and obtain the joint region and edge probability maps from the last convolutional layer
\begin{equation}
(P_R,P_E)=\textbf{CNN}(I,\textbf{W}^*),
\end{equation}
where \textbf{CNN}($\cdot$) denotes the trained network. $P_R$ is a vector region probability map including $n$ (number of regions) channels, while $P_E$ is a scalar edge probability map. These probability maps are then fed to the energy (\ref{eq:transformedFunction}), in which $f_i = -{\rm{log}}P_{Ri}, i=\{1,...,n\}$ and $g=P_E$. With all necessary elements at hand, we are ready to minimise (\ref{eq:transformedFunction}) next. \\

\noindent \textbf{Optimisation:} The minimisation process of (\ref{eq:transformedFunction}) entails the \textit{calculus of variations}, by which we obtain the resulting Euler-Lagrange (EL) equation with respect to the variable $\phi$. A solution ($\phi^*$) to the EL equation is then iteratively sought by the following gradient descent method
\begin{equation} \label{eq:GD}
\frac{{\partial \phi }}{{\partial t}} =  - \sum\limits_{i = 1}^n {{f_i}\frac{{\partial {\chi _i}\phi }}{{\partial \phi }}}  + \lambda {\kappa _g}\sum\limits_{i = 1}^{n - 1} {\delta_\epsilon \left( {\phi  - {c_i}} \right)},
\end{equation}
where ${\kappa _g} = div\left( {g{{\nabla \phi } \mathord{\left/{\vphantom {{\nabla \phi } {\left| {\nabla \phi } \right|}}} \right.\kern-\nulldelimiterspace} {\left| {\nabla \phi } \right|}}} \right)$ is the weighted curvature that can be numerically implemented by the finite difference method on a half-point grid \cite{duan2015surface}. $\delta_\epsilon$ is the derivative of $H_\epsilon$, which is defined in \cite{chung2009image}. 

At steady state of (\ref{eq:GD}), a local or global minimiser of (\ref{eq:transformedFunction}) can be found. Note that the energy (\ref{eq:transformedFunction}) is nonconvex so it may have more than one global minimiser. To obtain a desirable segmentation result, we need a close initialisation of the level set function ($\phi^0$) such that the algorithm converges to the solution we want. We tackle this problem by thresholding the region probability map $P_{R3}$ and then computing the signed distance function (SDF) from the binary image using the fast sweeping algorithm. The resulting SDF is then used as $\phi^0$ for (\ref{eq:GD}). In this way, the whole optimisation process is fully automated. 

%% file: sections/results.tex

\section{Experimental results}
\label{sec:experiments}
\textbf{Data:} Experiments were performed using short-axis CMR images from 430 PH patients. For each patient 10 to 16 short-axis slices were acquired roughly covering the whole heart. Each short-axis image has resolution of $1.5 \times 1.5 \times 8.0\;\rm{mm}^3$. Due to the large slice thickness of the short-axis slices and the inter-slice shift caused by respiratory motion, we train the FCN in a 2D fashion and apply the DNLS method to segment each slice separately. The ground truth region labels were generated using a semi-automatic process which included a manual correction step by an experienced clinical expert. Region labels for each subject contain the left and right ventricular blood pools and myocardial walls for all 430 subjects at end-diastolic (ED) and end-systolic (ES) frames. The ground truth edge labels are derived from the region label maps by identifying pixels with label transitions. The dataset was randomly split into training datasets (400 subjects) and validation datasets (30 subjects). For image pre-processing, all training images were reshaped to the same size of $160 \times 160$ with zero-padding, and image intensity was normalised to the range of $[0, 1]$ before training. 

 \textbf{Parameters:} The following parameters were used for the experiments in this work: First, there are six parameters associated with finding a desirable solution to (\ref{eq:transformedFunction}). They are the weighting parameter $\lambda$ (1), regularisation parameter $\epsilon$ (1.5), two levels $c_1$ (0) and $c_2$ (8), time step $t$ (0.1), and iteration number (200). Second, for training the network, we use Adam SGD with learning rate (0.001) and batch size (two subjects) for each of 50000 iterations. The weight $\alpha$ in (\ref{eq:REloss}) is set to 1. We perform data augmentation on-the-fly, which includes random translation, rotation, scaling and intensity rescaling of the input images and labels at each iteration. In this way, the network is robust against new images as it has seen millions of different inputs by the end of training. Note that data augmentation is crucial to obtain better results. Training took approx. 10 hours (50000 iterations) on a Nvidia Titan XP GPU, while testing took 5s in order to segment all the images for one subject at ED and ES. 
\begin{figure}[h!] 
\vspace{-20pt}
\centering  
{\includegraphics[width=0.99\textwidth]{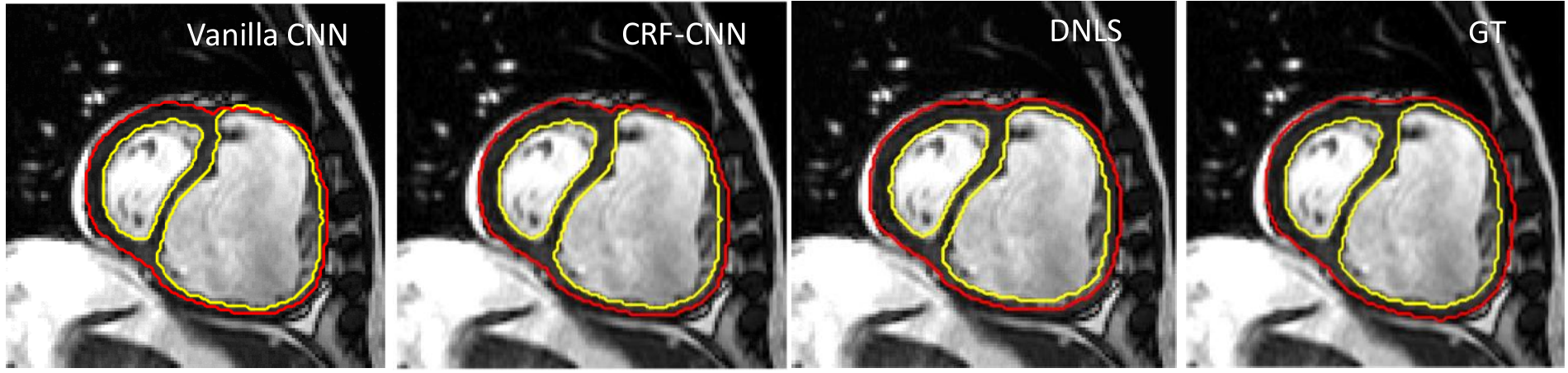}}\\
\vspace{-5pt}
\caption{Visual comparison of segmentation results from the vanilla CNN, CRF-CNN and proposed method. LV$\&$RV cavities and myocardium are delineated using yellow and red contours. GT stands for ground truth.}
\label{fig:visual}
\vspace{-25pt}
\end{figure}

\begin{table}[h!]
\vspace{-25pt}
\centering
\caption{Quantitative comparison of segmentation results from the vanilla CNN, CRF-CNN and proposed method, in terms of Dice metric (mean$\pm$standard deviation) and computation time at testing stage.}
\begin{tabular}{ccccc} \toprule
{Methods} & {LV\&RV Cavities}     & {Myocardium} & {Time} \\ \midrule
Vanilla CNN \cite{bai2017human}            & 0.902$\pm$0.047          & 0.703$\pm$0.091           & $\sim$\textbf{0.06}s  \\
CRF-CNN     \cite{krahenbuhl2011efficient} & 0.911$\pm$0.045          & 0.712$\pm$0.082           & $\sim$2s  \\
Proposed DNLS                              &\textbf{0.925$\pm$0.032} & \textbf{0.772$\pm$0.058}  & $\sim$5s  \\ \midrule
\end{tabular}
\label{tb:numb}
\vspace{-15pt}
\end{table}

 \textbf{Comparsion:} The segmentation performance was evaluated by computing the Dice overlap metric between the automated and ground truth segmentations for LV $\&$ RV cavities and myocardium. We compared our method with the vanilla CNN proposed in \cite{bai2017human}, the code of which is publicly available. DNLS was also compared with the vanilla CNN with a conditional random field (CRF) \cite{krahenbuhl2011efficient} refinement (CRF-CNN). In Fig~\ref{fig:visual}, visual comparison suggests that DNLS provides significant segmentation improvements over CNN and CRF-CNN. For example, at the base of the right ventricle both CNN and CRF-CNN fail to retain the correct anatomical relationship between endocardium and epicardium portraying the endocardial border outside the epicardium. CRF-CNN by contrast retains the endocardial border within the epicardium, as described in the ground truth. In Table~\ref{tb:numb}, we report their Dice metric of ED and ES time frames in the validation dataset and show that our DNLS method outperforms the other two methods for all the anatomical structures, especially for the myocardium. CNN is the fastest method as it was deployed with GPU, and DNLS is the most computationally expensive method due to its complex optimisation processes.

%% file: sections/conclusion.tex

\section{Conclusion}
\label{sec:conclusion}
In this paper, we proposed the deep nested level set (DNLS) approach for segmentation of CMR images in patients with pulmonary hypertension. The main contribution is that we combined the classical level set method with the prevalent fully convolutional network to address the problem of pathological image segmentation, which is a major challenge in medical image segmentation. The DNLS inherits advantages of both level set method and neural network, the former being able to model complex geometries of cardiac morphology and the latter providing robust features. We have shown the derivation of DNLS in detail and demonstrated that DNLS outperforms two state-of-the-art methods. 